\documentclass[10pt,letterpaper,twocolumn]{article}
\usepackage[margin=1in]{geometry}
\usepackage{times}
\usepackage{helvet}
\usepackage{courier}
\usepackage{graphicx}
\usepackage{amsmath, amsthm}
\usepackage{natbib}
\usepackage{caption}
\usepackage{natbib}
\usepackage{titlesec}
\titleformat{\section}{\large\bfseries}{\thesection}{0.5em}{}
\titlespacing*{\section}{0pt}{1em}{0.6ex}
\usepackage{amsthm}
\usepackage[hyphens]{url}
\usepackage{xurl}

\newtheorem{observation}{Observation}
\newtheorem{hypothesis}{Hypothesis}
\begin{document}

\twocolumn[

\begin{center}
{\large \bf Designing User-Centric Metrics for Evaluation of Counterfactual Explanations \par}
\vspace{1em}

{\normalsize \bf
Firdaus Ahmed Choudhury, Ethan Leicht, Jude Ethan Bislig, \par Hangzhi Guo, Amulya Yadav \par
}

\vspace{0.3em}

{\small
The Pennsylvania State University \par
University Park, PA 16802, USA \par
\{firdaus, ebl532, jeb6596, hangz, amulya\}@psu.edu
}
\vspace{1em}
\end{center}

]
\setcitestyle{authoryear,open={(},close={)}}

\begin{center}
\textbf{Abstract}
\end{center}

% Machine learning-based decision models are increasingly being used to make decisions that significantly impact people's lives, but their opaque nature leaves end users without a clear understanding of why a decision was made.
Counterfactual Explanations (CFEs) have grown in popularity as a means of offering actionable guidance by identifying the minimum changes in feature values required to flip an ML model's prediction to something more desirable. Unfortunately, most prior research in CFEs relies on \emph{artificial} evaluation metrics, such as proximity, which may overlook end-user preferences and constraints, e.g., the user's perception of effort needed to make certain feature changes may differ from that of the model designer. To address this research gap, this paper makes three novel contributions. First, we conduct a pilot study with 20 crowd-workers on Amazon MTurk to experimentally validate the alignment of existing CF evaluation metrics with real-world user preferences. Results show that user-preferred CFEs matched those based on proximity in only 63.81\% of cases, highlighting the limited applicability of these metrics in real-world settings. Second, inspired by the need to design a user-informed evaluation metric for CFEs, we conduct a more detailed two-day user study with 41 participants facing realistic credit application scenarios to find experimental support for or against three intuitive hypotheses that may explain how end users evaluate CFEs. Third, based on the findings of this second study, we propose the \emph{AWP} model, a novel user-centric, two-stage model that describes one possible mechanism by which users evaluate and select CFEs. Our results show that AWP predicts user-preferred CFEs with 84.37\% accuracy. Our study provides the first human-centered validation for personalized cost models in CFE generation and highlights the need for adaptive, user-centered evaluation metrics.

\section{Introduction}
%Advances in Machine Learning (ML) have led to the % [rapid] 
%deployment of predictive models in various aspects of human life, from approving credit applications \citep{faheem-2021} to job recruitment \citep{alAlawi-naureen-alAlawi-AlHadad-2021}. However, these advancements come at the cost of transparency, as complex black-box ML models provide little to no insight about how they arrive at their decisions \citep{herse-et-al-2018}. As such, there is a growing demand not only from end users \citep{binns2018} but also from regulatory frameworks like the EU's General Data Protection Regulation (GDPR) for increased transparency and accountability in AI/ML based decision support systems \citep{goodman2016eu}.

%To this end, 
Counterfactual explanations (CFEs) have emerged as a promising approach for providing intuitive, actionable insights into the decision-making process of black-box ML models \citep{wachter-2018}. More formally, CFEs provide post-hoc transparency for predictions returned by a black-box ML model on an input point, $x$, by identifying the minimal changes to feature values needed to convert $x$ to a new artificially constructed data point, $x'$, that achieves a more desirable prediction outcome from the ML model. The recent rise in popularity of CFEs is due to their potential to provide actionable recourse to end users who have received adverse decisions from automated ML-based decision systems \citep{karimi-scholkopf-valera-2021}. For example, a loan applicant denied credit by a bank's ML model might be shown that increasing their income by \$4,000 would get their application approved. %These counterfactual explanations can help users approximate and interpret the model's decision-making process by providing an intuitive human-readable rationale for how specific feature changes can alter model predictions \citep{spreitzer-EvaluatPracticCF-Italy-2022}. 

% CFEs are being increasingly used to provide actionable recourse to end users who have received adverse decisions from automated ML-based decision systems \citep{karimi-scholkopf-valera-2021}. Although CFEs have numerous benefits, the quality of recourses hinge on the evaluation metrics used to generate them. Technical metrics such as proximity, sparsity, and feasibility strive to create CFs with the least amount of change that still remain feasible under the model's constraints
Unfortunately, existing CFE algorithms rely on artificial metrics such as proximity, sparsity, etc., to optimize and evaluate generated recourses\footnote{Counterfactual explanations and algorithmic recourse are used interchangeably in the rest of the paper} \citep{wachter-2018}.
% More critically, many existing algorithms for generating CFEs directly optimize for these metrics, particularly proximity \citep{wachter-2018} being the most commonly optimized metric in existing CFE generation algorithms.
There is a growing consensus that these metrics ignore human preferences and the effort required to implement a proposed change \citep{CHOU202259, verma-et-al-2024}. E.g., a CF recourse recommending an applicant to get a master's degree to secure a loan, when they currently only hold a high school diploma, may score well on proximity, but is impractical in reality \citep{guo-counternet-2023, mahajan-tan-sharma-2020}. Similarly, \citet{tominaga-yamashita-kurashima-2024} also found that users rated counterfactuals optimized for proximity and sparsity poorly, with users describing them as unrealistic or undesirable.

%\begin{figure}[t]
%\centering
%\includegraphics[width=0.95\columnwidth]{graphs/CF.pdf}
%\caption{ \textbf{Counterfactual Generation.} This diagram illustrates how CFEs are generated by identifying alternative data points ($x'$) near the original input ($x$) that would cross the model’s decision boundary and lead to a different prediction. several counterfactual candidates ($x'$) relative to an original input ($x$) and a model’s decision boundary. Some candidates have low proximity to the original point but are invalid, while others are valid but require a larger change, highlighting the trade-off between minimizing distance and ensuring validity.}
%\label{fig_cf}
%\end{figure}

Despite this consensus, there has been very little empirical validation of these metrics with real users \citep{brughmans2024nice}. 
% In fact, \citet{keane2021five} reports that only 21\% of existing studies included user evaluations for specific counterfactual methods, with just 7\% conducting comparative user studies to benchmark different approaches. In a separate line of work, 
\citet{rawal-lakkaraju-BT-model-2024} proposed learning individualized feature costs using pairwise comparisons of recourses, though their work remains untested with real users.
This lack of rigorous human validation highlights a fundamental gap in CFE research: without testing methods with real users, we cannot know if they truly reflect the needs of the end user. As a result, the disconnect between how researchers evaluate CFEs and how users interpret or act on them continues to grow, potentially leading to CFEs that are technically sound but practically irrelevant.

To address this critical research gap, we make three novel contributions. First, we conduct a pilot study with 20 crowd-workers on Amazon Mechanical Turk to experimentally validate the alignment of existing CF evaluation metrics with the real-world preferences of end users. 
% Results show that user-preferred CFEs matched those based on proximity and sparsity in only 63.81\% and 69.51\% of cases, respectively.
% , highlighting the limited applicability of these metrics in real-world settings. 
% Second, inspired by the need to design a user-inspired evaluation metric for CFEs,
Second, we conduct a more detailed two-day user study with 41 participants facing realistic credit application scenarios to test three intuitive hypotheses about how end users evaluate CFEs. We capture both quantitative preferences and qualitative reasoning of participants, to get a deeper understanding of the factors they consider when choosing between recourses. Finally, based on our results from this second user study, we propose the \textbf{A}cceptability \& \textbf{W}eighted \textbf{P}roximity (AWP) model, a novel two-stage user-centric framework that describes one possible mechanism used by users to evaluate and choose CFEs. 

\section{Related Work}\label{sec:relatedwork}
\noindent \textbf{Counterfactual Explanations.} A counterfactual method aims to find a minimally different point x' from input x, such that the model prediction changes (validity) \citep{wachter-2018}, while minimizing the distance between the two points (proximity) using a particular distance metric, typically the L1 norm \citep{karimi-et-al-2020}

 % A counterfactual method for data point $x$ in model $f(.)$ aims to find a minimally different point $x'$, where (1) the distance between $x$ and $x'$ is minimized on a particular distance metric, typically the $L_{1}$ norm \citep{karimi-et-al-2020} (proximity objective in CFE algorithms), and (2) their model predictions are different, i.e., $f(x)\neq{}f(x')$ \citep{wachter-2018} (validity objective in CFE algorithms).
 % As CFEs identify the smallest set of changes needed to alter a prediction, they naturally align with the goals of algorithmic recourse: to help users understand the cause behind unfavorable decisions they received from a predictive model and subsequently provide them with actionable steps they can take to overturn those decisions. \citep{ustun-spangher-liu-2019,tominaga-yamashita-kurashima-2024}.

 % \noindent \textbf{Algorithms for CFE Generation.} 
 \citet{wachter-2018} introduced an optimization-based method that jointly optimizes CFEs on a combined loss of validity and proximity via gradient descent. Subsequent methods have built on their work and emerged with additional objectives, such as diversity \citep{mothilal-sharma-tan-2020}, feasibility \citep{mahajan-tan-sharma-2020}, and sparsity \citep{keane2021five}. Unfortunately, most algorithms rely on artificial loss functions and evaluation metrics to find the optimal CFE.
 
% \subsection{Evaluation Metrics for Recourse Quality}
% Wachter's proposed algorithm optimized on the combined loss of prediction change and proximity to input distance via gradient descent. Building on this foundational framework, subsequent methods have come up with additional objectives beyond proximity, such as diversity, generating multiple distinct counterfactuals \citep{mothilal-sharma-tan-2020}; feasibility, ensuring changes are realistic and satisfy causal relations between features \citep{mahajan-tan-sharma-2020}; and sparsity, minimizing the number of feature changes \citep{keane2021five}.
However, recent work has shown that artificial metrics fail to account for user preferences \citep{tominaga-yamashita-kurashima-2024}. As a result, there is growing recognition that generating realistic algorithmic recourses requires evaluation metrics to go beyond model-centric methods and incorporate user-centered perspectives \citep{verma-et-al-2024}. 
% In contrast, our work evaluates counterfactuals using a weighted proximity metric, where weights, learned through a Bradley-Terry model, represent the user's perceived costs of changing specific features \citep{rawal-lakkaraju-BT-model-2024}. Incorporating personal costs allows us to generate recourses that better reflect each user's unique preferences.

\noindent \textbf{Generating Human-Centric CFEs. }
\citet{mahajan-tan-sharma-2020} introduced constraints based on causal graphs to generate valid recourses and briefly suggested learning user-specific constraints by treating users as oracles, though this idea remained unimplemented \& untested.  \citet{rawal-lakkaraju-BT-model-2024} proposed learning individual feature costs using pairwise comparisons via the Bradley-Terry model \citep{bradley-terry-1952}, though their method was only tested in simulation. Our work builds on this and provides the first real user empirical validation.
% GAM Coach \citep{wang2023gam} allows users to edit feature preferences and difficulty levels but depends on a glass-box model. The authors themselves acknowledge that this transparency raises many security concerns, such as users gaming the system or being victim to model inversion attacks \citep{xu-et-al-2025}. Due to its limited applicability to black-box models, we do not consider it as a baseline.
UFCE \citep{suffian-et-al-2024} uses binary feedback to optimize a generator via Mutual Information, but captures only general patterns of recourse preference. Although it allows constraints on feature changes, it does not learn feature-specific costs. In contrast, our approach uses pairwise comparisons to model user-specific preferences.

\noindent \textbf{User Centric Evaluation of Algorithmic Recourses}
A related but distinct body of research emphasizes the importance of evaluating CFEs with real users. \citet{keane2021five} critically examined gaps in the evaluation of counterfactual methods, and found that only 21\% of studies included user evaluations, with just 7\% conducting comparative user studies to benchmark different approaches. %Their meta-analysis echoes a growing recognition in similar work \citep{verma-et-al-2024} that human validation is essential for assessing the usefulness and realism of algorithmic recourses.  
In response to this gap, recent studies have focused on user-centered evaluations in counterfactual research. For instance, \citet{warren2023categorical} conducted two large-scale experiments with over 330 participants, finding that CFEs improved predictive accuracy over causal or no explanations. They also showed that CFEs with categorical changes were easier to interpret than mixed-type ones. \citet{tominaga-yamashita-kurashima-2024} demonstrated, through a user study, that CFEs optimizing for proximity and sparsity often misalign with user preferences, reinforcing the argument that strong results on technical benchmarks do not always reflect real-world utility. 
% Our work advances this area by conducting user evaluations that not only validate a proposed counterfactual method, but also uncover how users reason about and choose between alternative recourses.

\section{Methodology}
\label{sec:method}
% This section outlines our methodology for generating and evaluating counterfactual recourses, which serves as a foundation for running our user studies.
% in which we: (i) explain our choice of using a decision tree classifier (trained on a synthetic credit application dataset) as the ML model for which CFEs need to be generated; followed by (ii) discussing how CFEs are generated for this decision tree classifier. (iii), we introduce the distance metrics used to quantify recourse proximity, including standard and personalized formulations. (iv) Finally, 
 % First, we outline our approach to recourse generation, then we discuss how feature-level weights are learned from participant responses using the Bradley-Terry model. These weights are integrated into our weighted proximity metric to generate personalized recourses, which are later compared against traditional proximity-based recourses using user evaluations.
% \subsection{Recourse Generation}
Our process of generating counterfactual recourses follows a structured pipeline involving (1) synthetic data creation, (2) loan decision modeling, (3) recourse generation, and (4) learning preference-based weighting for individual users.

\subsection{Synthetic Dataset Creation}
We constructed a synthetic dataset simulating binary loan application outcomes (approved vs. rejected) to maintain greater experimental control. Although several real-world credit datasets exist \citep{statlog_(german_credit_data)_144, fico2018}, they often include high-dimensional, noisy, or domain-specific features that can introduce confounding factors and increase the cognitive load on participants. Since our primary objective was to evaluate how lay users perceive and compare counterfactual recourses, a simplified feature space allowed participants to focus on the core trade-offs without being overwhelmed by extraneous details.

Each data point represents a loan application with five continuous and categorical features commonly used in lending: \emph{income}, \emph{credit score}, \emph{employment status}, \emph{education level}, and the \emph{loan amount} requested. Feature values were independently sampled from predefined, realistic ranges that reflect plausible applicant profiles. 
% These data points were labeled using a rule-based scoring function modeled on typical lending criteria.
To simulate real-world variability, we implemented multiple rule sets, each capturing a different combination of feature thresholds and decision logic for loan approval or rejection. Full details of sampling and labeling procedures are provided in the Appendix. 
% In addition, we publicly release a link to our synthetic dataset for full reproducibility.

\subsection{Training the Prediction Model}
We trained a decision tree classifier on our synthetic dataset to predict loan approval outcomes. This model served as the basis for generating CFEs, which were later presented to participants.

While complex models such as neural networks might offer higher predictive accuracy, we deliberately chose decision trees for their transparency, interpretability, and rule-based structure. These characteristics made them a better fit for our study, where the goal was to evaluate how users perceive and reason about counterfactual recourses. In contrast, black-box models would have required the use of sophisticated, often opaque CFE generation algorithms. While such models might have produced technically stronger recourses, they would have significantly limited our ability to control the types of counterfactuals shown to participants and made it more difficult to study user preferences about CFEs.

The decision tree model was trained using a standard 80/20 train-test split of the synthetic dataset and achieved 99.15\% accuracy on the test set, confirming that the rule sets effectively captured the feature-label relationships. Details of our training procedure are provided in the Appendix. 
% Additionally, all our code is publicly released for full reproducibility.
%Unlike most black-box ML models, decision trees provide an explicit mapping between input features and model prediction outcomes, allowing us to trace the path from a rejection leaf to an acceptance leaf and verify whether a counterfactual followed a truly minimal path in terms of feature changes. This transparency was critical for identifying the features that influenced the decision boundary and helped ensure that small feature changes led to understandable outcome flips. %Since the primary objective of our study was to evaluate how humans respond to different counterfactual generation strategies, not to optimize predictive performance, we prioritized model interpretability and experimental control. This made decision tree classifiers an ideal choice for our experimental framework.
 \subsection{Counterfactual Generation} Unlike most black-box ML models, decision trees 
 % provide an explicit rule-based mapping between input features and model prediction outcomes, which  allow us to trace the exact \emph{decision path} for any given data point, i.e.,
 allow us to trace the data point's sequence of feature-based splits from the root to the leaf node. Let $L$ denote a leaf node in our decision tree, and let $\phi(L) \in \{\text{accept}, \text{reject}\}$ denote the label assigned to all data points that reach $L$.

\begin{figure}[t]
\centering
\includegraphics[width=0.9\columnwidth]{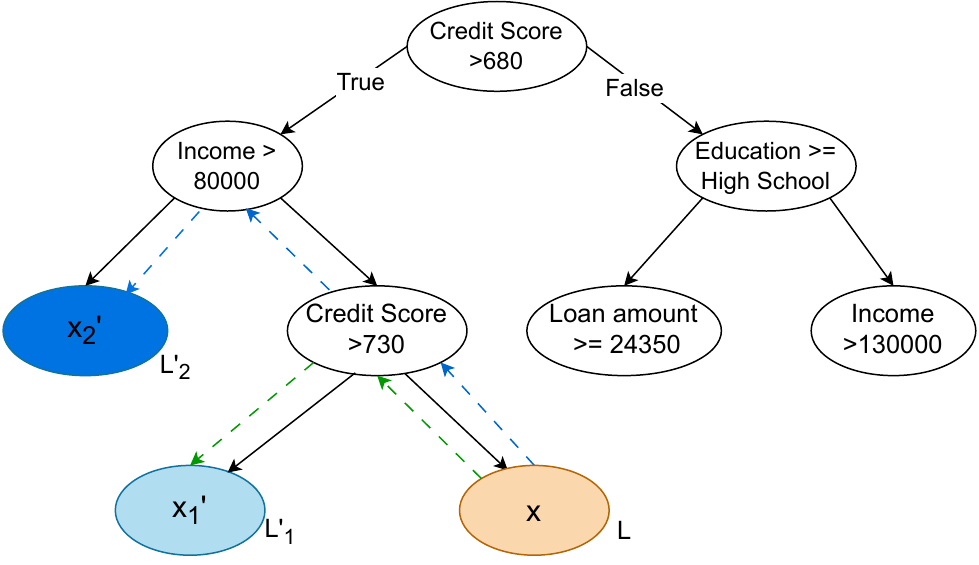}
\caption{ \textbf{CFE generation with a Decision Tree Classifier.} Let $x$ denote a rejected loan application that reaches the red leaf $L$ in the decision tree. In order to generate counterfactual recourses $x'_1$ and $x'_2$ (that reside in leaves $L'_1$ and $L'_2$, respectively), the shortest path between leaves $L$ and $L'_1$ (or $L'_2$) needs to be traversed. In this figure, these shortest paths are denoted using dotted green and blue paths, respectively.}
\label{fig_decisionTree}
\end{figure}

Our strategy for generating CFEs for our decision tree classifier (illustrated in Fig.\ref{fig_decisionTree}) leverages a key structural insight: for each rejected loan application $x$ that reaches leaf node $L$ (with $\phi(L) = \text{reject}$), we identify counterfactuals by finding paths to nearby leaf nodes $L'$ where $\phi(L') = \text{accept}$. Each path corresponds to a set of feature changes required to shift the decision tree's prediction from rejection to approval. For example, in Fig.\ref{fig_decisionTree}, a counterfactual example $x'_1$ is constructed for input $x$ by applying the feature modifications suggested along the path between $x$ and $x'_1$. This way, we formulate CFE generation for decision trees as a structured search over the tree, where nearby oppositely labeled leaf nodes are identified by enumerating candidate target leaves and comparing their decision paths with that of the original instance to extract the minimal set of differing decision constraints.
%the path from leaf nodes associated with a rejection label to leaf nodes associated with an acceptance label, and verify whether this ``counterfactual" followed a truly minimal path in terms of feature changes. This transparency was critical for identifying the features that influenced the decision boundary and helped ensure that small feature changes led to understandable outcome flips.

Using this simple rule-based algorithm, we generate CFEs for all test set instances classified as 'rejected' by the loan prediction model. For each 'rejected' instance $x$, we extract the top-15 CFEs corresponding to distinct accepting leaf node $L'_i$ (i.e., $\phi(L'_i) = \text{accept}$ for all $i \in {1, \ldots, K}$). These 15 CFEs are ranked in increasing order of proximity to $x$.

To ensure realism and interpretability of generated recourses, we discard CFEs that suggest implausible or counterintuitive transformations. For example, continuous features such as \emph{income} and \emph{credit score} are only allowed to increase in a valid recourse, reflecting feasible, forward-looking improvements in creditworthiness. For each point $x$, we continue generating candidate CFEs until we identify $K$ such realistic and semantically meaningful CFEs.

\noindent \textbf{Evaluating Generated Counterfactuals. }To evaluate the quality of CFEs generated in the previous step, we employ two primary evaluation metrics:
\begin{itemize}
\item 
\textit{\bf{Proximity}} is central to most CFE generation algorithms \citep{wachter-2018, karimi-et-al-2020}. Proximity is measured as the cumulative cost of changes required to convert input data $x$ to counterfactual $x'$. It is most commonly computed as the normalized $L_1$ distance between $x$ and $x'$. 
\[
   \text{\bf{Prox}}(x,x')= \sum_{i=1}^{d} \frac{|x_i - x_i'|}{\text{Range}_i}
   \]
% ,i.e., Prox(x,x')= $\frac{|x_i - x_i'|}{\text{Range}_i}$.
% \[
% $\text{\bf{Prox}(x,x')}= \sum_{i=1}^{d} w_i\frac{|x_i - x_i'|}{\text{Range}_i}$.
% \]
\item
\textit{\bf{Weighted Proximity}} is calculated by assigning a distinct feature weight $w_i$ to feature $i$ that reflects its perceived cost of modification to that user. Larger values of $w_i$ correspond to a greater perceived difficulty in modifying that feature value. It calculates the weighted cost of change required to convert input $x$ to CFE $x'$. We operationalize this as the weighted $L_1$ norm between $x$ and $x'$.
% , i.e., $\text{WeightedProx}(x, x') = \sum_{i=1}^{d} w_i \cdot \frac{|x_i - x_i'|}{\text{Range}_i}$
   % $\text{\bf{WeightedProx}(x, x')} = \sum_{i=1}^{d} w_i \cdot \frac{|x_i - x_i'|}{\text{Range}_i}$
   \[
   \text{\bf{WeightedProx}}(x, x') = \sum_{i=1}^{d} w_i \cdot \frac{|x_i - x_i'|}{\text{Range}_i}
   \]
\end{itemize}

The standard proximity metric assumes all feature changes are equally feasible; for example, modifying income is as easy as adjusting the loan amount. In practice, users perceive the difficulty of changing features differently. By incorporating user-specific feature weights, the weighted proximity metric offers a more personalized and realistic assessment of recourse feasibility, aligning evaluation with each user's individual preferences and constraints.\\

\subsection{Learning User-Specific Feature Weights}
To learn individual feature weights $w_i$ for a user, \citet{rawal-lakkaraju-BT-model-2024} propose using pairwise comparison data, where each data point corresponds to a pair of possible CFEs, $x'_1$ and $x'_2$ (for input point $x$), which are presented to a user who is then asked to choose which they prefer. Based on these comparisons, the authors fit a Bradley-Terry model, a probabilistic framework that estimates a strength parameter $\beta$ for each feature by modeling the likelihood of one feature being preferred over another \citep{bradley-terry-1952}. They show that the additive inverse of the learned $\beta_i$ values can be used to represent individual feature-specific weights ($w_i = -\beta_i$). 

% Using simulated pairwise comparison data, \citet{rawal-lakkaraju-BT-model-2024} show to fit a Bradley-Terry model, which is a probabilistic framework used to model the likelihood of pairwise outcomes by estimating a strength parameter $\beta$ for each item (feature) in a set based on the observed outcomes \citep{bradley-terry-1952}. Finally, \citet{rawal-lakkaraju-BT-model-2024} show that the additive complement of the inferred $\beta_i$ parameters (in the learned Bradley-Terry model) can be used to accurately represent individual user-specific feature weights, i.e., $w_i = -\beta_i$. 

In our work, we collect pairwise comparison data from each study participant and fit individual Bradley-Terry models  (utilizing the design of \citet{rawal-lakkaraju-BT-model-2024}). As a result, we can learn personalized feature weights that reflect each participant's preferences, allowing for the generation of CFEs tailored to their willingness to act.

\section{User Validation of CFE Metrics}\label{sec:pilot}
As discussed in Section~\ref{sec:relatedwork}, the field of CFEs has developed mainly around mathematically convenient, but often artificial, evaluation metrics such as proximity and sparsity. These metrics have rarely been subjected to systematic validation from the perspective of real users. 

To the best of our knowledge, the only prior work that empirically examines user acceptance of CFEs is done by  \citet{tominaga-yamashita-kurashima-2024}. Their findings suggest that users’ willingness to accept a recourse does not correlate with their proximity/sparsity score, casting doubt on the foundational assumptions behind these metrics. However, their study does not investigate more personalized evaluation functions, which may better capture the diverse constraints and preferences of users.

\noindent \textbf{Pilot Study Design. }To address this gap, we conducted an initial pilot user study to investigate: \emph{Are users more likely to accept CFEs that optimize standard proximity (or sparsity) vs those that optimize weighted proximity?} To explore this, we designed a survey containing 20 CFE scenarios. Each scenario presented a rejected loan application alongside two alternative CFE recourses. The survey was administered to 20 participants recruited via Amazon Mechanical Turk (MTurk), each compensated \$3. Participants received a brief explanation of CFEs and were then directed to the Qualtrics survey. All user study protocols were approved by an Institutional Review Board at the host institution.

Scenarios were selected through a multi-step process. As described in Section \ref{sec:method}, we used our shortest-path algorithm to generate a top-$15$ set of CFEs ${x'_1, x'_2, \ldots, x'_{15}}$ for each rejected instance $x$ in the test set. Next, we examined all possible CFE pairs $(x'_i, x'_j)$ from this set and included a pair if, without loss of generality, $x'_i$ had a lower proximity score than $x'_j$, while $x'_j$ had a lower weighted proximity score than $x'_i$, or vice versa. Weighted proximity was computed using a fixed set of author-defined weights reflecting the perceived difficulty of modifying each feature. 
% Pairs not satisfying this divergence were excluded from consideration.
We randomly selected 20 such qualifying CFE scenarios to populate the survey. A sample survey question is provided in the Appendix (Fig.\ref{fig_pilot}).

For each of the CFE scenarios presented to a pilot study participant, they were asked to select which recourse (out of $x'_i$ and $x'_j$) was more acceptable to them. Additionally, they were asked to give a brief reason for why they chose this recourse over the other. %At the end of the scenario-based tasks, participants were asked to rank the features of the credit rejection scenario from most to least willing to change in order to get a loan approved.

% \begin{figure}[t]
% \centering
% \includegraphics[width=0.9\columnwidth]{graphs/pilot_1.png}
% \caption{ \textbf{Screenshot of a question from the pilot study.} The participant is shown two alternative profiles (Profile 1 and Profile 2) with modified applicant attributes that would lead to loan approval, prompting them to choose one and explain their reasoning.}
% \label{fig_day2}
% \end{figure}

% \begin{figure}[t]
% \centering
% \includegraphics[width=0.9\columnwidth]{graphs/pilot_2.png}
% \caption{ \textbf{Screenshot of a question from the pilot study.} The participant is shown two alternative profiles (Profile 1 and Profile 2) with modified applicant attributes that would lead to loan approval, prompting them to choose one and explain their reasoning.}
% \label{fig_day2}
% \end{figure}

\noindent \textbf{Pilot Study Results. }On average, participants selected the recourse with lower weighted proximity in only 36.29\% of the cases.

\begin{observation}
    This low selection rate raises concern about the suitability of weighted proximity when computed using fixed global weights as a proxy for user preferences.
\end{observation}

 However, when we analyzed the open-ended explanations provided by participants, it became evident that their aversion to low-weighted proximity recourses was driven by a mismatch between their perceived cost of feature changes and the global weights we had set. For example, p6 expressed that they "would prefer to make a change in employment type (from Private sector to Government sector) rather than reducing the amount [they are] requesting," a view not aligned with our global weights.

\begin{observation}
    Fixed feature weights fail to capture individual users' subjective notions of the feasibility of change, leading to a misalignment between user preferences.
\end{observation}

Different participants showed a large variation in their preference rankings. For example, p7 ranked their preference of changing features as loan amount $<$ credit score $<$ income $<$ employment type $<$ education, whereas p20 ranked their preference as loan amount $<$ income $<$ education $<$ employment type $<$ credit score.

\begin{observation}
    Participants exhibited highly individualized notions of feature change difficulty, suggesting the need for personalized feature weights. These qualitative findings seem to suggest that people's acceptability of weighted proximity as a metric should only be measured when personalized feature costs are learned for individual users.
\end{observation}

Despite the fact that we probably used weights that were not aligned with each user's personalized weights, standard proximity-based recourses were only chosen in about 63\% of the cases. Similarly, standard sparsity-based recourses were only chosen in about 69\% of the cases.

\begin{observation}
    Unweighted proximity and sparsity fail to sufficiently capture user preferences.
\end{observation}

In addition, participants' explanations for their choice of preferred recourse revealed another interesting observation: in several cases, participants selected a recourse not because it was particularly desirable, but because the alternative was deemed unacceptable - typically due to it requiring the participant to modify feature values to implausibly large levels. 

\begin{observation}
    Users are willing to consider a recourse as acceptable as long as its suggested feature modifications remain below some threshold.
\end{observation}

\section{Two-Day User Study to Understand User Preferences}
Our pilot study revealed a notable misalignment between widely used evaluation metrics and actual user preferences. Even in a simplified setting, only 63\% and 69\% of participants preferred CFEs optimized for raw proximity and sparsity over those based on weighted proximity using global feature weights. These results highlight the limitations of current CFE evaluation metrics in accurately capturing what users truly value, underscoring the need for more realistic, user-centered evaluation metrics for CFE generation.

In this paper, we take a step in this direction by proposing AWP, a two-stage user-centric model that can accurately predict which CFE would be most preferred by an end user. As shown in our experimental results in Section \ref{sec:experiments}, AWP improves alignment with user preferences by 34\% over existing baseline metrics like proximity, and achieves an 84.37\% accuracy at predicting user-preferred CFEs.

However, to design a truly user-centric AWP evaluation model, it is crucial to systematically understand and analyze the first-order principles that end users may use in choosing a preferred counterfactual. 
% A deeper understanding of these first-order principles can then be incorporated in the design and development of AWP. 
To this end, we conducted a detailed two-part semi-structured interview study with 43 participants. The primary goal of this study was to evaluate four research hypotheses, 
% each reflecting a distinct first-order principle that may influence user preferences over CFEs. These hypotheses were 
informed by both quantitative patterns and qualitative insights observed in our initial pilot study.

Observations 2 and 3 from Section~\ref{sec:pilot}, highlighted that users often disagreed with the fixed feature weights used during the pilot, suggesting the need for personalized feature weights. Observation 5 revealed that users frequently rejected CFEs when the proposed change to a feature exceeded what they personally considered an acceptable shift, indicating the presence of feature-specific acceptability thresholds. These observations motivated the following two hypotheses:

\begin{hypothesis}
Users prefer CFEs that minimize weighted proximity, where weights, $w_i$, are personalized to reflect the individual user's perceived difficulty in modifying feature $i$.
\end{hypothesis}

\begin{hypothesis}
Users possess a set of feature-specific acceptability thresholds ($\alpha_i$ for feature $i$); they are likely to reject any CF explanation that recommends changing the value of feature $i$ beyond their individual threshold $\alpha_i$.
\end{hypothesis}

Prior research in cognitive psychology \citep{tversky1974judgment} suggests that individuals often rely on heuristic processing, particularly when faced with complex or cognitively demanding decisions. 
% Drawing on this insight, we hypothesize that users may prefer CFEs with rounded feature values (e.g., income of \$5500) over those with precise but cognitively taxing values (e.g., income of \$5257.29), even when the latter offer objectively smaller changes (with better weighted proximity).
For instance, a user earning \$5000/month may find it more intuitive to increase their income to \$5500 rather than to an exact figure like \$5257.29. Although the latter results in a lower weighted proximity, the increased cognitive burden associated with processing and interpreting the unrounded feature value may make it less desirable. This intuition motivates our third hypothesis:

\begin{hypothesis}
Users prefer CFEs that involve rounded feature value changes over those that suggest highly precise, unrounded values, even when the latter result in a lower weighted proximity.
\end{hypothesis}

\begin{figure}[t]
\centering
\includegraphics[width=0.95\columnwidth]{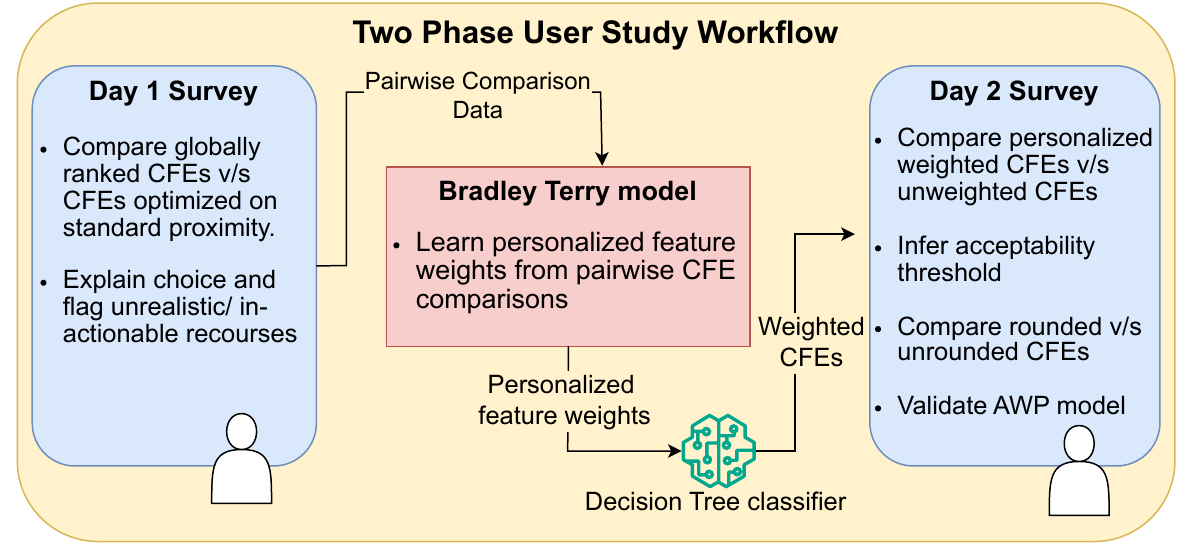}
\caption{ Workflow of our Two-Phase User Study}
\label{workflow}
\end{figure}

\noindent \textbf{User Study Design. }To validate our three hypotheses, we conducted an in-depth user study exploring how individuals evaluate and select counterfactual recourses. Each participant took part in two one-on-one Zoom interview sessions, approximately one hour each. This two-session format enabled us to learn personalized feature weights for each participant separately, which was necessary to validate Hypothesis 1. A schematic of the study workflow is shown in Fig.\ref{workflow}.
 
Participants were recruited using a combination of convenience and snowball sampling, including targeted emails to student mailing lists at a U.S. public research university, personal outreach, and social media posts. Additionally, we also encouraged participants to share the study invitation within their networks to increase reach and diversity. Eligibility was restricted to individuals aged 18 years or older and currently residing in the United States. A total of 43 participants were recruited, of whom 41 completed both sessions; two incomplete cases were excluded from further analysis. Each participant was given a \$30 Amazon gift card as compensation. %for the study, but only 41 completed both sessions; we excluded the two incomplete cases from our analysis.
%In total, we recruited 41 participants who completed the study.

Each of the two sessions was structured to serve a distinct purpose: the first session focused on eliciting participants' preferences through scenario-based pairwise comparisons of CFEs, which could then be used to learn personalized feature weights for each user separately. The second session focused on finding evidence for/against our three hypotheses by presenting users with CFEs tailored to their personalized feature weights. All sessions were conducted with informed consent, recorded with permission, and transcribed to facilitate qualitative analysis of user preferences.

%Each participant completed two video-based interview sessions, scheduled approximately one week apart. The first session focused on eliciting personalized preferences through scenario-based comparisons, while the second session evaluated the effectiveness of counterfactuals generated using personalized feature costs and aimed to learn additional factors participants consider when making decisions. Before beginning the study, we obtained informed consent from all participants. With their permission, we recorded each session and transcribed the audio to support qualitative analysis of their reasoning and reflections.
%Based on the insights we get from the user study, We want to explore:
%\begin{itemize}
%    \item How does incorporating feature-specific weights from the Bradley-Terry model affect how real humans choose recourses?
%    \item What decision strategies do users employ when comparing and selecting between a pair of recourses?
%    \item Does rounding continuous decimal features impact the interpretability and acceptance of recourses?
%\end{itemize}

\noindent \textbf{Session 1: Collecting Pairwise Comparison Data. } 
% The first session was designed to elicit personalized preferences for feature changes by asking participants
Participants were presented with 25 credit rejection scenarios and asked to imagine themselves in those situations. For each scenario, they had to choose the counterfactual recourse they would be more willing to act upon between two alternative recourses, to get their loan approved. 

Each scenario contained two CFEs that spanned a diverse set of feature change combinations, constructed to reflect a deliberate trade-off: one recourse had a lower raw (unweighted) proximity, while the other had a lower weighted proximity based on globally defined feature rankings derived from our pilot study. This design ensured that the two options emphasized different strengths, enabling us to isolate which evaluation metric was preferred by participants.
% , e.g., if counterfactual A had lower raw proximity, then counterfactual B had lower global weighted proximity, and vice versa. 

Using this comparison data, we employed the approach in \citet{rawal-lakkaraju-BT-model-2024} to fit a Bradley-Terry model for each participant, learning their personalized weights $w_i$ for each feature $i$. To complement this quantitative choice data, we also asked participants to explain the reasoning behind each of their choices, giving us qualitative insights into how they perceived feasibility, effort, and desirability.  %After completing all scenario-based questions, we asked participants to rank all individual features by their willingness to change them to receive a loan. We then used each participant’s pairwise choices to fit a Bradley–Terry model, which allowed us to learn a personalized feature cost function based on their observed preferences.

\begin{figure}[t]
\centering
\includegraphics[width=0.95\columnwidth]{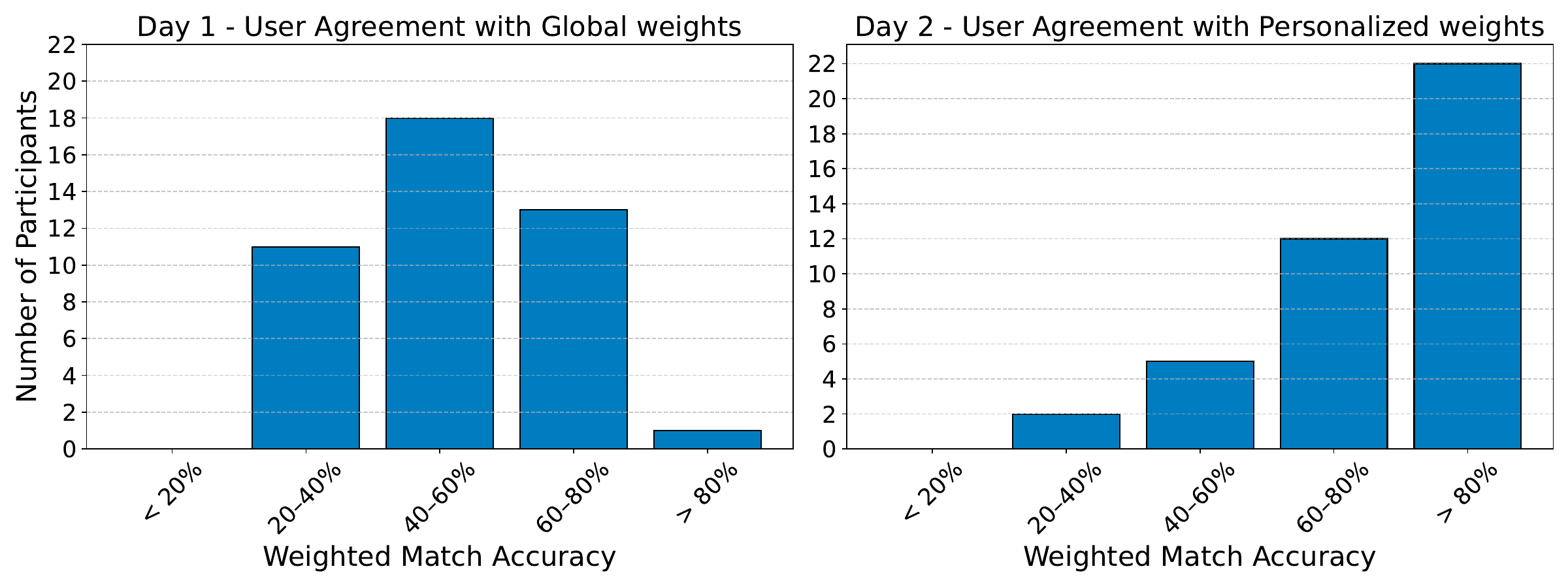}
\caption{ \textbf{Participant selection rates for recourses generated using weighted proximity with non-personalized weights (left) vs personalized weights (right).} The majority of participants aligned only moderately (40–60\%) with global-weighted recourses, whereas they aligned very strongly ($>$80\%) with personalized-weighted recourses.}
\label{fig_day1_2}
\end{figure}

\noindent \textbf{Session 2: Validating Hypotheses. }
Each of the 41 returning participants was presented with a personalized survey containing 34 credit rejection scenarios generated using their learned feature weights from Session 1.

To validate Hypothesis 1, we randomly sampled 15 rejected loan applications from our synthetic dataset. For each of these scenarios, we generated two CFEs: one optimized for raw proximity, while the other achieved a lower weighted proximity, leveraging user-specific weights derived from a Bradley-Terry model. Participants were asked to pick their preferred CFE in each scenario and provide a brief explanation justifying their choice. Care was taken to ensure that the two recourses in each loan scenario differed in at least one feature to ensure meaningful comparison.

\begin{figure}[t]
\centering
\includegraphics[width=0.99\columnwidth]{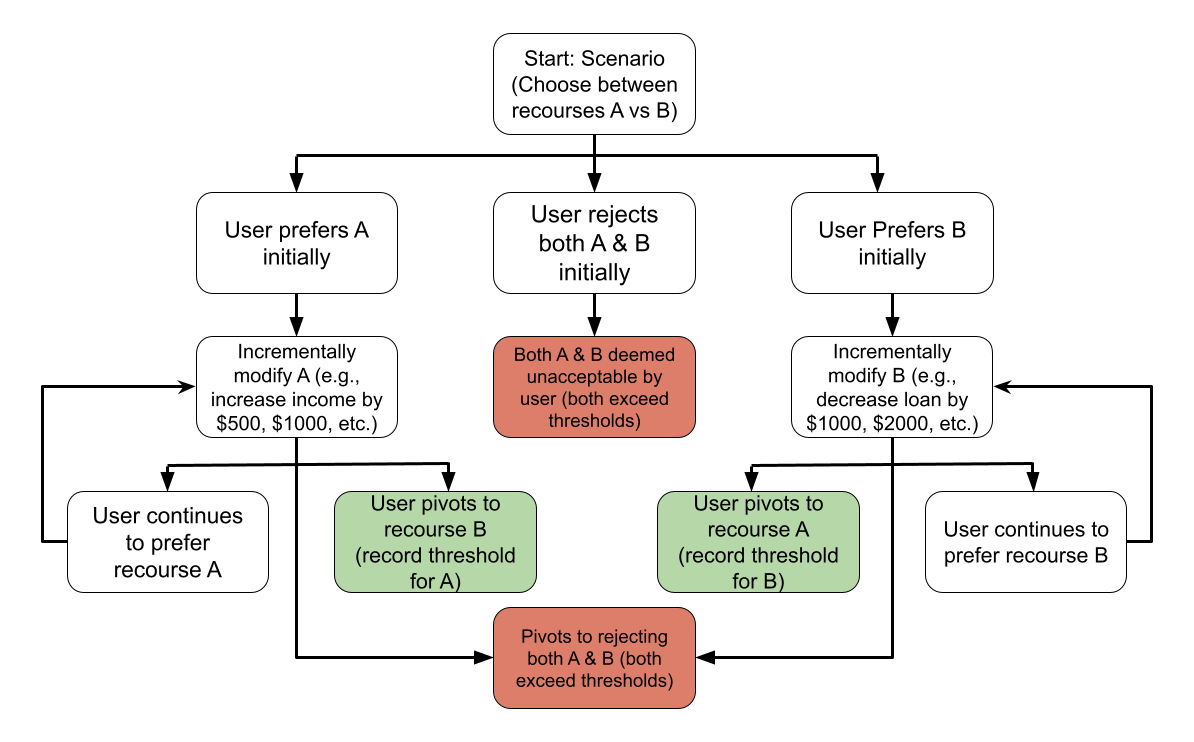}
\caption{\textbf{Flowchart describing our iterative probing strategy to validate Hypothesis 2.} Green cells correspond to decision points at which at least one (or more) feature-specific acceptability thresholds for the user can be inferred.}
\label{probing}
\end{figure}

To validate Hypothesis 2, we incorporated 9 additional CFE scenarios designed to probe participants’ feature-specific acceptability thresholds through one-on-one, interactive discussions. As shown in Fig.\ref{probing}, for each scenario, participants were first asked to choose between two distinct recourses. Once a preference was expressed, e.g., if a participant chose Recourse A, which required increasing income from \$5000 to \$5500 over Recourse B, which involved lowering the requested loan amount from \$10,000 to \$9000, the interviewer initiated a sequence of follow-up comparisons.

Recourse A was incrementally modified by increasing the magnitude of change in the same feature (e.g., income increased to \$6000, then \$6500, etc), while Recourse B remained unchanged. At each step, participants were asked whether they preferred the updated version of A (denoted A'), whether they would prefer B, or reject both options altogether. This iterative process continued until the participant switched preference or indicated that neither was acceptable.

These pivots were used to infer feature-specific acceptability thresholds. If a participant ultimately rejected both recourses, it indicated that A' surpassed their personal threshold and that B had never been within their acceptable range. The last instance in which A' was still preferred is interpreted as the user’s acceptability threshold for that particular feature. E.g., if a participant accepted A' up to an income change of \$8000 but rejected at \$8500, we infer that their income acceptability threshold lies between those values.

Alternatively, if the participant eventually preferred B, we reversed the process by incrementally modifying B in the same way and comparing B' with the last accepted A' (or A). We can identify an approximate acceptability threshold for the feature modified in Recourse B, based on the point at which the participant's preference shifted or their tolerance was exceeded. This two-sided probing procedure enabled us to establish the existence of individualized feature-specific thresholds for multiple attributes, capturing the amount of change a participant was willing to accept in a given feature. During this exercise, we also corroborated their reasons for pivoting to ensure that they were indeed pivoting because the amount of change recommended by a particular recourse exceeded their threshold of acceptability.

%We designed these surveys to evaluate how participants respond to standard counterfactual recourses when compared against personalized recourses generated using the feature weights learned from their Day 1 responses. The primary goal was to evaluate whether incorporating user-specific preferences results in recourses that are perceived as more actionable and desirable.

%Participants received a new set of credit rejection scenarios. For each scenario, they were shown two counterfactual recourses: one generated using a standard unweighted proximity metric, and the other using a weighted proximity metric based on the participant's feature weights derived from the Bradley-Terry model. The two recourses differed in at least one feature change to ensure meaningful comparison. Since the weighted recourses were personalized, each participant received a unique survey tailored to their individual preferences. We asked participants to select which of the two recourses they were more likely to act on and briefly explain the reasoning behind their choice. 
 
To validate Hypothesis 3, we included 10 CFE scenarios designed to assess the impact of potential cognitive overload on user preferences. In each scenario, participants were presented with two recourses that were functionally equivalent but differed in how those changes were expressed. One recourse used precisely calculated (unrounded) continuous feature values (e.g., increasing income from \$5000 USD to \$5257.29 USD), while the other showed a rounded version of the same change (e.g., to \$5500 USD). The direction of rounding was also varied in different scenarios, allowing us to isolate the effect of rounding on user preferences.

\noindent \textbf{User Study Results. }Are divided into three main findings:

\noindent \textit{\textbf{A. Preference for 'Personalized' Weighted Proximity:}}
On average, participants selected recourses optimized on their personalized feature weights 80.2\% of the time, demonstrating a strong preference for individualized counterfactuals over those optimized for raw proximity. As illustrated in Fig.\ref{fig_day1_2}, over half (22 out of 41) participants selected CFEs with lower weighted proximity in over 80\% of hyp 1 scenarios. Three-fourths (31 out of 41 participants) did so in at least 70\% of scenarios. Only two participants preferred unweighted recourses in the majority of the scenarios, choosing personalized ones in 40\% of the cases for hyp 1. 

This result provides strong empirical support for Hypothesis 1, demonstrating that personalized weighted proximity is closely aligned with actual user preferences. While researchers have discussed weighted proximity in CFEs for a long time, no existing work has focused on validating it with real users. To the best of our knowledge, our work is the first to provide 
% end-user-driven validation of the effectiveness of personalized weighted proximity as an evaluation metric for CFEs. It also provides the first-ever 
an empirical validation of the method proposed by \citet{rawal-lakkaraju-BT-model-2024} for learning feature-specific weights with real users.

\begin{figure}[t]
\centering
\includegraphics[width=0.95\columnwidth]{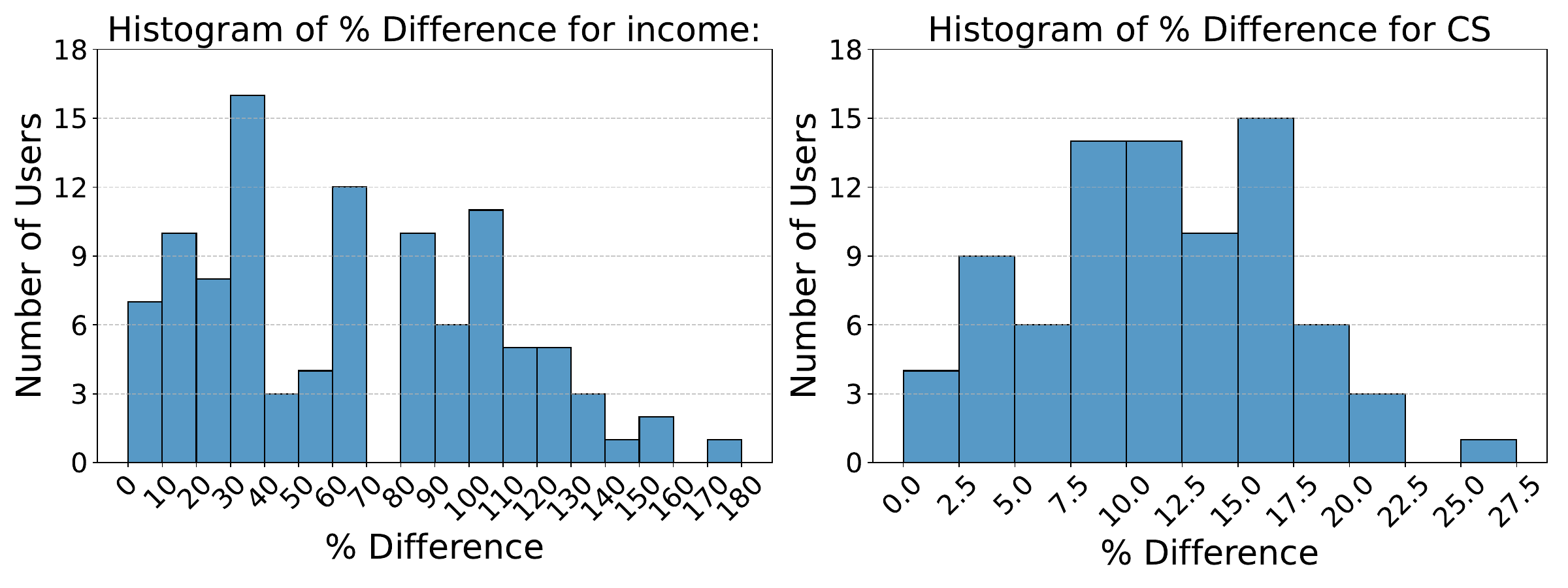}
\caption{Histogram of acceptability thresholds for \emph{income} and \emph{credit score} features inferred through our iterative probing scenario across all 41 participants.}
\label{fig_day2}
\end{figure}

\noindent \textit{\textbf{B. Strong Evidence for Acceptability Thresholds:}}
Our iterative probing revealed clear feature-specific thresholds beyond which participants rejected or pivoted away from a preferred CFE. These decisions were not arbitrary; participants explicitly cited feasibility concerns. For instance, P11 said that "[they] would not pick any credit score change above 780 because to cross that point would require years of on-time payments." P26 talked about how they "can't realistically achieve income increases more than 20\% because they would need to change their job to get that increase." Comments like these confirm that perceived feasibility, not just numerical distance, drives the acceptability of a recourse. Fig.\ref{fig_day2} shows the histogram of acceptability thresholds for \emph{income} and \emph{credit score} (Refer to Fig.\ref{fig_hyp3all} in the appendix for all features). These findings provide empirical evidence for feature-specific acceptability thresholds.

In real-world applications, these feature thresholds do not need to be learned; the user usually applies for a recourse from a known baseline and can specify their own limits, which can be incorporated as a hard constraint in the recourse generation process. This ensures that generated recourses align with the end user's acceptability thresholds. 

% \begin{figure}[t]
% \centering
% \includegraphics[width=0.9\columnwidth]{graphs/day2_cymk.pdf}
% \caption{ \textbf{Distribution of participant selection rates for recourses generated using personalized feature weights on Day 2.} The majority of participants selected personalized counterfactuals over 80\% of the time, demonstrating a strong alignment with their individualized feature costs.}
% \label{fig_day2}
% \end{figure}

\noindent \textit{\textbf{C. Limited Impact of Rounding: }}  On average, participants picked recourses with precise continuous values in 49.26\% of scenarios, and chose rounded values in 50.64\% of cases. 
%in an overwhelming number of cases. %in X\% scenarios.
When rounded recourses were chosen, participants' justifications made it clear that rounding itself was not the deciding factor; their choices were guided by the perceived ease of achieving the change or its potential long-term benefits (e.g., a higher credit score improving creditworthiness). A common theme among the participants was that they "prefer better numbers rather than nicer looking numbers", as said by P43. Our results suggest that either users did not experience sufficient cognitive overload with our 5-feature recourses or that users' cognitive overload was not sufficiently alleviated by our use of the rounding-based heuristic.

Our user study reveals strong evidence in support of Hypothesis 1 and 2, whereas we observe strong evidence against Hypothesis 3. As mentioned earlier, these hypotheses correspond to potential first-order principles that may be leveraged by end users in choosing their most preferred recourse. Building on these findings, we develop the AWP model, a framework that explains one possible mechanism used by participants in choosing a preferred counterfactual.

\section{Acceptability \& Weighted Proximity Model}\label{sec:experiments}
The Acceptability \& Weighted Proximity (AWP) model takes as input four things: (i) the original data point which gets an unfavorable prediction; (ii) a set of alternative recourses, each suggesting a distinct set of feature changes that results in a more favorable model prediction; (iii) personalized feature-specific weights $w_i$ for that user; and (iv) feature-specific acceptability thresholds $\alpha_i$ for that user. Based on these inputs, the AWP model predicts which provided recourse would be most preferred by the user.

The AWP model utilizes the following two-stage decision process when selecting between counterfactual recourses:
\noindent \textbf{Stage 1 – Feasibility Filtering: } For each of the provided recourses, the end user first assesses whether the changes suggested by that recourse are "acceptable" or not. This is done by ensuring that all feature changes are within the user's corresponding acceptability threshold $\alpha_i$. If a certain change exceeds $\alpha_i$, the whole recourse is deemed "unacceptable" by the user and is filtered out from further analysis. Note that this first stage in the AWP model is motivated by the strong empirical evidence found in favor of Hypothesis 2.

\noindent \textbf{Stage 2 – Weighted Comparison: } Among the recourses that passed the feasibility filtering, users select the recourse with the lowest personalized weighted proximity, calculated using user-specific feature weights $w_i$. Note that this second stage in the AWP model is motivated by the strong empirical evidence found in favor of Hypothesis 1.

\noindent \textbf{Predictive Accuracy Results with AWP Model. } To evaluate AWP's predictive accuracy in determining user preferences, we conducted a retrospective analysis using data from our user study (with 41 participants). CFE scenarios presented during Session 2 for Hypothesis 2 were categorized into three bins: (i) scenarios where participants rejected both recourses from the outset - indicating that both recourses were deemed unacceptable in Stage 1 of AWP; (ii) scenarios where participants initially preferred one recourse but later rejected both options after incrementally modifying the original option - cases where the preferred recourse was acceptable, however, the alternative was unacceptable and filtered out; (iii) scenarios in which participants initially preferred one recourse, but after threshold for acceptability was breached, they pivoted to preferring the alternative recourse. 

To compute the predictive accuracy of the AWP model, we restricted our evaluation to the third category, as it represents cases where both alternative recourses were within the user's acceptability thresholds, passing Stage 1 of the AWP model. In the first and second categories, respectively, there were either no acceptable recourses or a single acceptable recourse available for stage 2 of AWP. In either of these cases, the user's choice in Stage 2 would be trivial. 

For each scenario in the third category, we assessed whether the participant selected the recourse with the lower personalized weighted proximity, calculated using the feature weights inferred for that participant from their Session 1 data.
Across a total of 369 CFE scenarios presented during Session 2, 32 scenarios contained alternative recourses that were both acceptable to the user. In 27
% 84.4/%
of these cases, participants chose the recourse with lower personalized weighted proximity, yielding a predictive accuracy of 84.4\% for AWP.

To our knowledge, AWP is a first-of-its-kind model for predicting user preferences by evaluating the quality of CFEs. AWP has been derived using first-order principles that are validated using extensive human participant data. However, we acknowledge that since AWP was derived from the same study data used in this evaluation, this result may partially reflect overfitting to the observed behavioral patterns. A more rigorous, validation on independent data will be necessary to confirm the generalizability of AWP as a user preference model—an important direction for future work.

\section{Conclusion}
% Counterfactual Explanations have grown in popularity as a means of offering actionable recourse in a wide variety of domains. Unfortunately,
% Most prior research in CFE relies on \emph{artificial} evaluation metrics, such as proximity, which may potentially overlook end-user preferences and constraints. To address this research gap, this paper makes three novel contributions. First, we conduct a pilot study with 20 crowd-workers on Amazon Mechanical Turk to experimentally validate the alignment of existing CF evaluation metrics with the real-world preferences of end users. Results show that user-preferred CFEs matched those based on proximity in only 63.81\% of cases, highlighting the limited applicability of these metrics in real-world settings. Second, inspired by the need to design a user-inspired evaluation metric for CFEs, we conduct a more detailed two-day user study with 41 participants facing realistic credit application scenarios to find experimental support for/against three intuitive hypotheses that may explain how end users evaluate CFEs. Third, based on our results from this second user study, we propose the Acceptability \& Weighted Proximity (AWP) model, a novel two-stage user-centric model that describes one possible mechanism used by end users to evaluate and choose CFEs of their choice. Our experimental results show that \emph{AWP} predicts user-preferred CFEs with 84.37\% accuracy. Our study provides the first human-centered validation for personalized cost models in CFE generation and highlights the need for adaptive, user-centered evaluation metrics.

Our work questions the prevailing reliance on artificial metrics, such as proximity and sparsity, in generating counterfactual explanations. We conduct two empirical studies, which show that these metrics often overlook end-user preferences and constraints. We uncover clear evidence that user preferences are shaped by personalized perceptions of feature difficulty and individual thresholds of acceptability. These insights led us to propose the AWP model, a novel two-stage user-centric model describing one possible mechanism used by end users to evaluate and choose CFEs of their choice. Importantly, our study provides the first human-centered validation of personalized cost models in CFE generation. Our results call for a broader rethinking of evaluation metrics in explainable AI, shifting from a one-size-fits-all formulation to adaptive, user-informed systems that account for individuality. While our study offers promising results, it requires validation across more diverse populations. Secondly, our study focuses only on recourses for credit rejection scenarios, restricting the applicability of our results to a single domain. Preferences may differ across different decision-making tasks. Future work should test the generalizability of AWP across domains, expand it to more complex decision-making contexts, and explore integrations with real-world decision support systems.

\bibliography{references}

\appendix

\clearpage

\section{Synthetic Dataset generation}
We began by synthesizing a dataset full of loan application profiles. We decided upon five basic features that would be included in each loan profile. To ensure the profiles we generated had some resemblance to reality in the United States, the distribution of values was carefully selected for each feature.

In 2023, the median personal income in the United States was approximately \$42,000, and most of the total wealth is held by a small percentage of people. The log-normal distribution with a mean of \$42,000 and a standard deviation of 0.8 provides a good approximation of the income distribution with an accurate median and mean. Incomes were excluded below \$10,000 and above \$500,000 to avoid extreme outliers.

\begin{figure*}[t]
\centering
\includegraphics[width=0.95\textwidth]{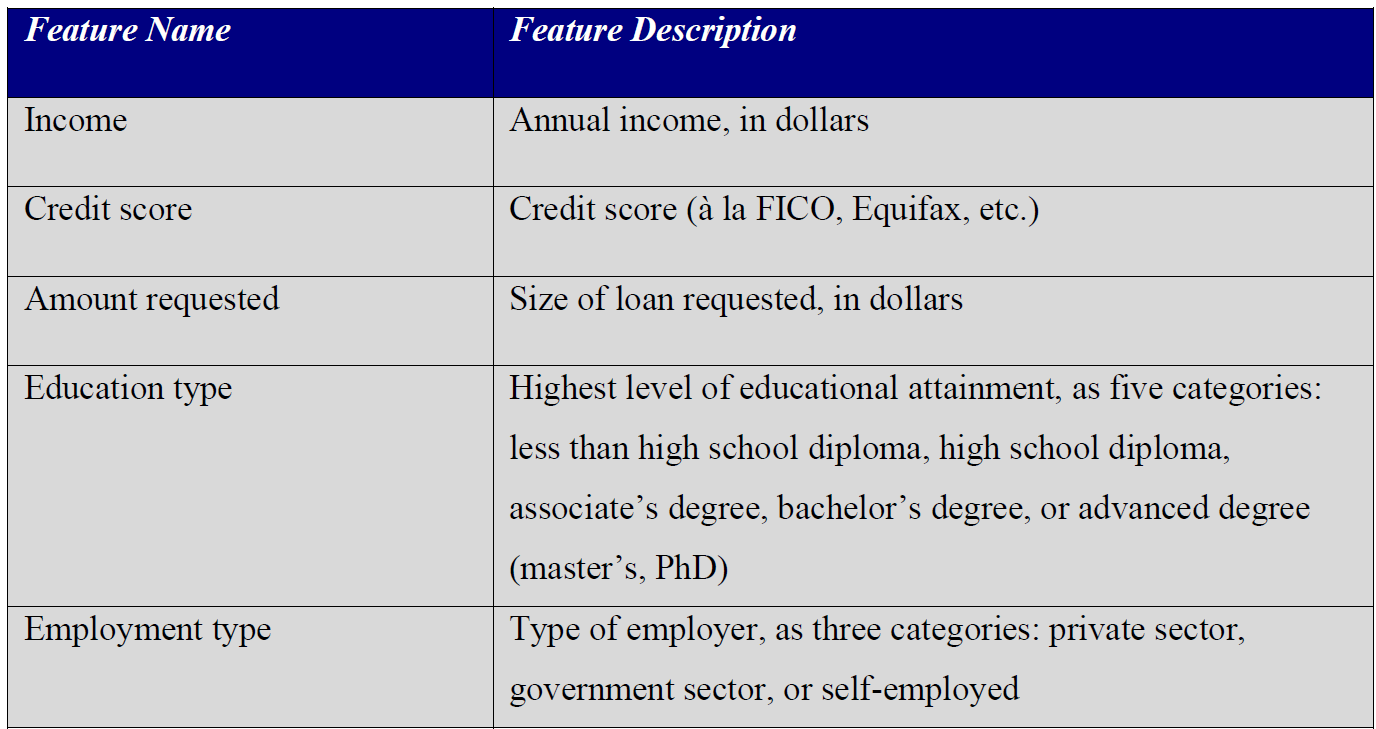}
\caption{ \textbf{Loan profile feature descriptions} }
\label{fig_loan_feat_desc}
\end{figure*}

Credit scores are integers and fall in the range of 300 to 850. The average credit score in the United States is approximately 700. A beta distribution with parameters a=2 and b=4 created the appropriate left-skewed distribution

For simplicity, a uniform distribution was used to generate loan sizes. The minimum loan size was set at \$1,000 and the maximum at \$50,000, which is the typical range permitted by most 
banks for personal loans.
Education type, or highest level of education, was divided as follows, based on recent United States educational attainment data.

Employment type was divided as follows, based on recent data from the Bureau of Labor Statistics.

100,000 unique profiles were generated by randomly sampling a value for each of the five features from the appropriate distribution. It should be noted that our dataset is not entirely realistic. We did not ensure different combinations of feature values was proportionally 
representative of the American population, nor did we consider the type of person who is more likely to seek a personal loan as a source of bias. Nonetheless, our data mapped onto reality well enough for the purposes of our study.

% \begin{links}
% \link{Code}{https://github.com/Anonymoussubmission855558/HumanCentricCFs}
% \end{links}

\section{Bradley-Terry Model}
The Bradley-Terry model is a probabilistic framework used to model the likelihood of pairwise outcomes by estimating a strength parameter $\beta$ for each item in a set based on the observed outcomes \citep{rawal-lakkaraju-BT-model-2024, bradley-terry-1952}. Given two items $i$ and $j$ belonging to a set of items, the model estimates the probability that item $i$ is preferred over item $j$ as:\[P_{i > j} = \frac{e^{\beta_i}}{e^{\beta_i} + e^{\beta_j}}\]
Here, $\beta_i$ and $\beta_j$ are the respective strength parameters of $i$ and $j$. A higher $\beta$ value corresponds to a stronger preference for that item.

To estimate the $\beta$ values from observed data, the model relies on collecting pairwise comparisons across all items in the set. Once a sufficient number of comparisons are collected, an empirical probability of preferring item $i$ over item $j$ is estimated by calculating the ratio of the times $i$ was chosen over $j$ to the total number of times $i$ and $j$ were compared. \[ \hat{P}_{i > j} = \frac{\#(i > j)}{\#(i > j) + \#(j > i)}\] These calculated probabilities serve as input for inferring the underlying strength parameters. We use Maximum A Posteriori (MAP) estimation, assuming a Gaussian prior, to determine the strength parameters \cite {rawal-lakkaraju-BT-model-2024}. Applying a zero-mean Gaussian prior ensures we get robust strengths even if certain features are never chosen \citep{fageot-et-al-2024}. 

In our setting, each item represents the individual features, $f$, that may be modified to generate a recourse. By collecting participants' pairwise preferences over which recourses they would prefer to implement, we are able to isolate the individual feature comparisons embedded within these choices. The comparisons are then aggregated into a feature-level win matrix to count how often one feature is preferred over another. The strength parameter for each feature is estimated using a simple iterative algorithm, applying MAP estimation. The resulting $\beta$ values capture the relative desirability of modifying each feature from the user's perspective \citep{rawal-lakkaraju-BT-model-2024}. Finally, we take the negative of the strength parameters as the perceived cost of changing that feature and normalize them to add up to 1. 
\[ Cost(f) = -\beta_{f}\]

\section{Loan Decision}
The next step was to develop an algorithm that our fictional bank will use to decide which profiles should be approved for their loan and which should be denied. The result was a 
multi-step approach.

First, income and amount requested were transformed into a new feature called
income:amount-requested, which is simply the individual’s income divided by their credit score, or the ratio between the two features. This new feature was more concise, as it replaced two features with one. It represents how much money the user is requesting relative to how much they make, which is something a bank is likely to consider in its decision.

The two continuous features, credit score and income:amount-requested were then normalized to a range of 0 to 1 using the min-max formula. Let x be a single value for a feature 
and x be the entire feature vector.

The two categorical features, employment type and education type, were mapped to numerical values in the range of 0 to 1, representing each category’s desirability score from the perspective of the bank. A higher desirability score means the category is more favorable for loan approval.

The rationale here is mostly intuitive. Higher levels of education are assigned higher desirability scores. An individual who is more educated is more likely to pay back their loan. Employment types are assigned desirability scores based upon their level of job security. An individual at higher risk of termination or business failure is less likely to pay back their loan.

With all values falling in a normalized range, we next assigned each of the four features a weight. The higher a feature’s weight, the more consequential it is to the bank’s decision. The weights are all positive and sum to 1.

Much like the categorical features’ desirability scores, feature weights were determined intuitively. When granting loans, banks are generally most concerned with an applicant’s credit 
score, as the very purpose of the metric is to indicate creditworthiness. The applicant’s income relative to their loan request is of the next highest import, as it suggests whether they can pay back what they take out. Education type and employment type are secondary factors that provide a snapshot of the applicant’s background, work ethic, and job security.

For every profile, each feature value was multiplied by the feature’s weight and summed together to obtain the profile’s overall desirability score. The higher a profile’s desirability score, the more likely the bank deems it that the applicant will pay back their loan.

Finally, all 100,000 loan profiles were sorted in ascending order of desirability score and assigned a binary digit. The first 50,000 were assigned a zero for denial, and the remaining 50,000 were assigned a one for approval. Each decision digit was then appended to the original, unmodified profile, with no additional information regarding how the decision was made.

\begin{figure*}[t]
\centering
\includegraphics[width=0.95\textwidth]{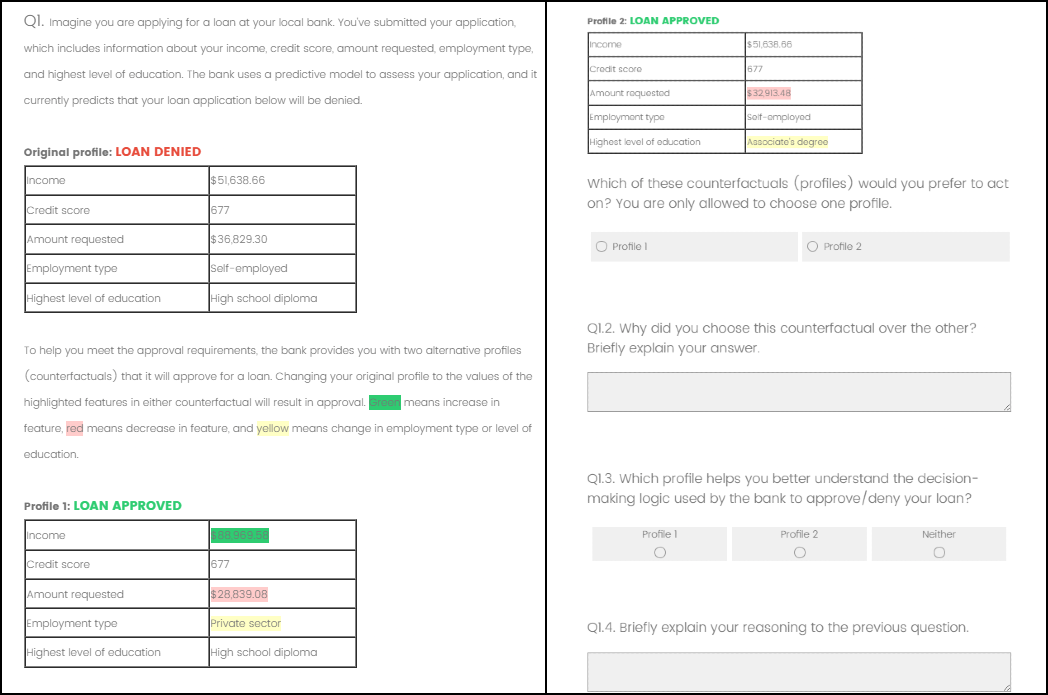}
\caption{ \textbf{Screenshot of a question from the pilot study.} The participant is shown two alternative profiles (Profile 1 and Profile 2) with modified applicant attributes that would lead to loan approval, prompting them to choose one and explain their reasoning.}
\label{fig_pilot}
\end{figure*}

\begin{figure*}[t]
\centering
\includegraphics[width=0.95\textwidth]{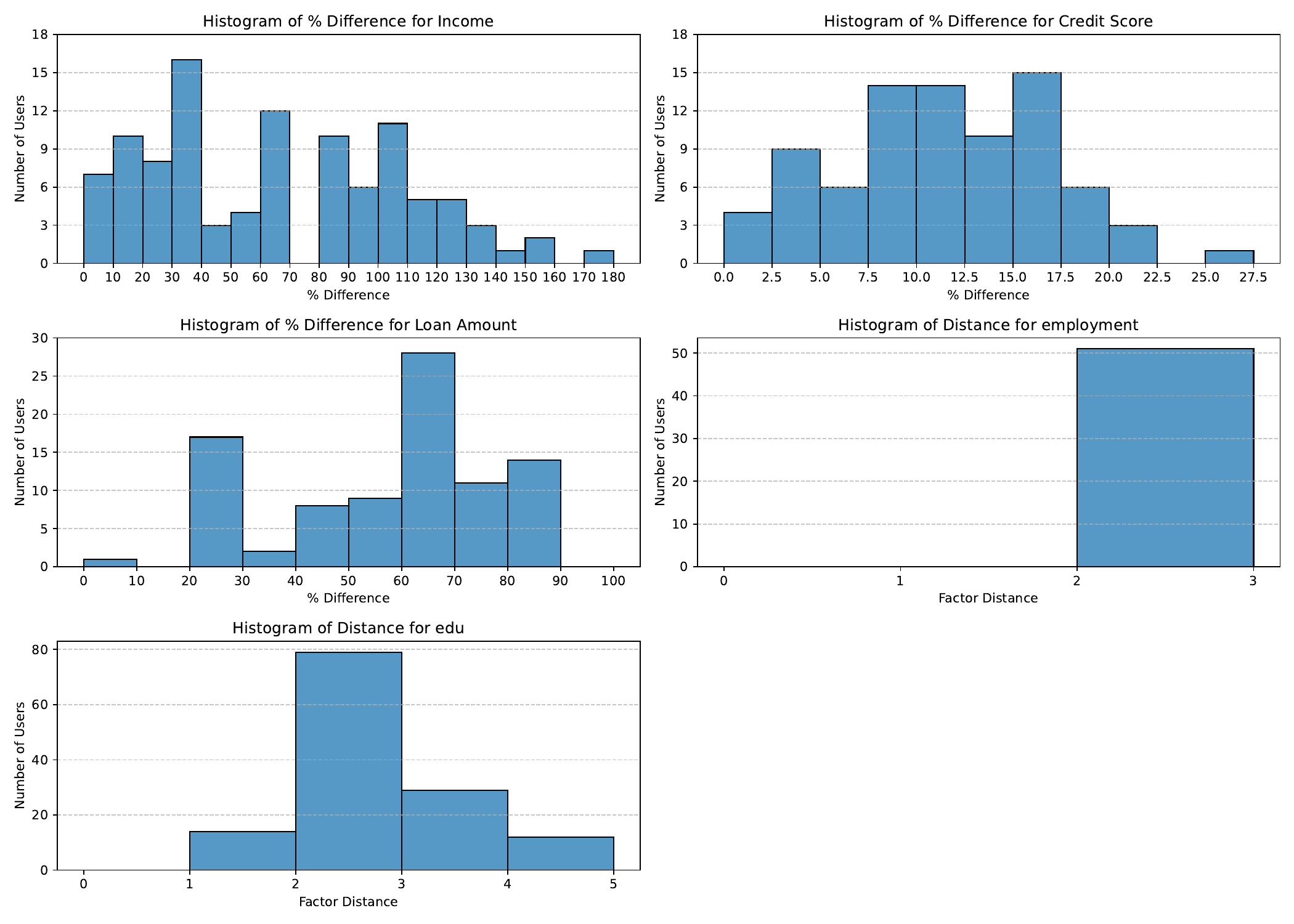}
\caption{Histogram of acceptability thresholds for all features inferred through our iterative probing scenario across all 41 participants.}
\label{fig_hyp3all}
\end{figure*}

\end{document}